\documentclass{article}
\usepackage{arxiv}
\usepackage[utf8]{inputenc}
\usepackage{graphicx}
\graphicspath{ {./images/} }
\usepackage{booktabs}
\usepackage{caption} % subfigurez
\usepackage{subcaption}
\usepackage{algorithm}
\usepackage{algorithmic}
\usepackage{amsmath}
\usepackage[dvipsnames]{xcolor}
\usepackage{comment}

\title{Accelerating Neural Architecture Exploration Across Modalities Using Genetic Algorithms}

\author{
    Daniel Cummings \\
    \small{Intel Labs, Intel Corporation} \\
    \small{daniel.cummings@intel.com} \\\And
    Sharath Nittur Sridhar \\
    \small{Intel Labs, Intel Corporation} \\
    \small{sharath.nittur.sridhar@intel.com} \\\And
    Anthony Sarah \\
    \small{Intel Labs, Intel Corporation} \\
    \small{anthony.sarah@intel.com} \\\And
    Maciej Szankin \\
    \small{Intel Labs, Intel Corporation} \\
    \small{maciej.szankin@intel.com}
}

\date{}

\begin{document}

\maketitle

\begin{abstract}
Neural architecture search (NAS), the study of automating the discovery of optimal deep neural network architectures for tasks in domains such as computer vision and natural language processing, has seen rapid growth in the machine learning research community. While there have been many recent advancements in NAS, there is still a significant focus on reducing the computational cost incurred when validating discovered architectures by making search more efficient. Evolutionary algorithms, specifically genetic algorithms, have a history of usage in NAS and continue to gain popularity versus other optimization approaches as a highly efficient way to explore the architecture objective space. Most NAS research efforts have centered around computer vision tasks and only recently have other modalities, such as the rapidly growing field of natural language processing, been investigated in depth. In this work, we show how genetic algorithms can be paired with lightly trained objective predictors in an iterative cycle to accelerate multi-objective architectural exploration in a way that works in the modalities of both machine translation and image classification. 
\end{abstract}

\section{Introduction}

Automating the process of finding optimal deep neural network (DNN) architectures for a given task, known as neural architecture search (NAS), has seen significant progress in the research community particularly in computer vision. However, the computational overhead of evaluating the performance of the discovered DNN architectures can be very costly due to the training and validation cycle. To address the training overhead, novel weight sharing approaches known as one-shot or super-networks \cite{darts2018, bender18a, bootstrapNAS} have offered a way to mitigate the training overhead by reducing training times from thousands to a few GPU days \cite{elsken2019neural}. These approaches train a task specific super-network architecture with a weight-sharing mechanism that allows the sub-networks to be treated as their own architectures. This enables sub-network model validation without a separate training cycle. However, the validation component still comes with a high overhead since there are many possible models to search across for large super-networks and the validation step itself comes with a computational cost, especially for larger datasets such as ImageNet \cite{imagenet}. One popular way to mitigate the validation cost is to train predictors for objectives such as inference time (a.k.a. latency) and accuracy from a training set with thousands of sampled architectures. Then the trained predictors are used for approximating model performance during the NAS process where only the best performing candidates are validated in the end. While reinforcement learning, sequential model-based optimization, and gradient optimization have been applied to the model search problem, evolutionary approaches, specifically genetic algorithms (GAs), have seen ongoing popularity in NAS work \cite{ren2021comprehensive}. GAs have been broadly applied specifically for computer vision NAS problems in both single-objective implementations \cite{guo2020single} and in multi-objective approaches such as NSGA-Net \cite{lu2019nsganet}. In the context of super-networks, they have been used to either inform a fine-tuned training of the super-network or to search sub-network architectures after training \cite{cai2020onceforall}. 

Most NAS research efforts have centered around the computer vision task of image classification and only recently have other modalities, such as the rapidly growing field of language modeling or language translation, been investigated in detail \cite{wang2020hat, feng21}. Moreover, understanding how NAS approaches generalize and perform across modalities and tasks has not been studied in depth. In this work we demonstrate how pairing GAs in an iterated fashion with lightly trained predictors can yield an accelerated and less costly exploration of the architecture search space in what we term as Lightweight Iterative NAS (LINAS). We show how this approach is extensible to machine translation NAS, given that most research has been focused on computer vision tasks and focus our NAS experiments on super-network frameworks given their large architectural design spaces. For the machine translation task, we use a transformer super-network architecture that has a search space size of $10^{15}$ \cite{wang2020hat}. For the image classification task, we apply our approach to a MobileNetV3 super-network that has a search space size of $10^{19}$ \cite{cai2020onceforall}. While many NAS approaches focus only on a single optimization objective, such as maximizing model accuracy for a particular latency or model complexity metric, we perform our experiments in the multi-objective context since searching for architectures that are optimized to a specific hardware platform (e.g., finding an optimal Pareto front for trade-offs in accuracy and latency) continues to be an important industry-wide application. 

\section{Proposed Algorithm}

\begin{figure}[htb]
    \centering
    \includegraphics[width=0.5\linewidth]{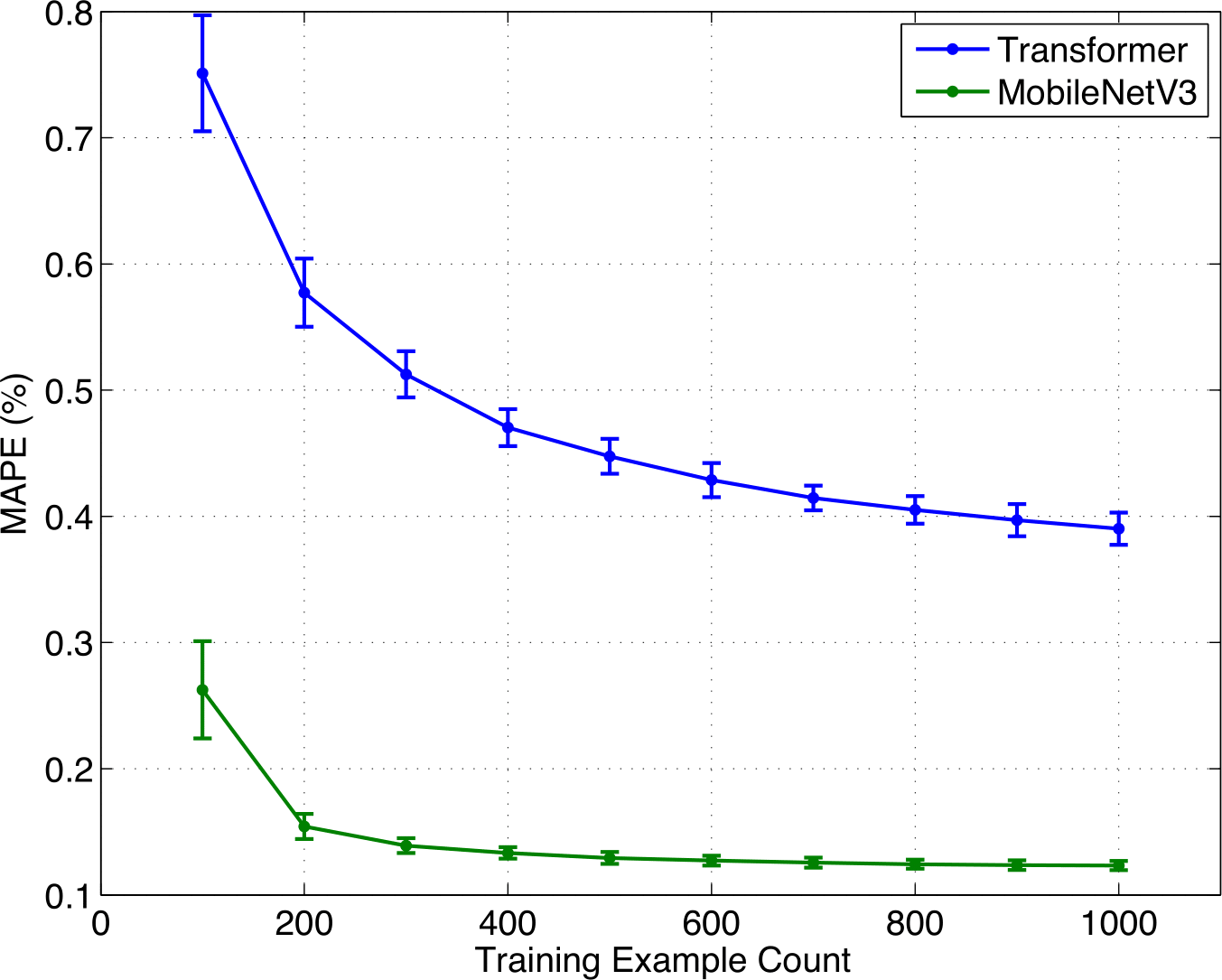}
    \caption{MAPE versus training examples for Transformer super-network BLEU score (SVR predictor) and MobileNetV3 super-network top-1 accuracy (ridge predictor). Each point is the average over 10 trials with error bars showing one standard deviation.} 
    \label{fig:predictors}
\end{figure}

The goal for our algorithm is to reduce the number of validation measurements that are required to find optimal DNN architectures in a multi-objective search space in a way that works well across modalities, in our case machine translation (e.g., Transformer super-network) and image classification (e.g., MobileNetV3 super-network). While existing work shows that using trained predictors can speed up the DNN architecture search process, there remains a substantial cost to training predictors since the number of validated training samples can often range between 1000 and 16000 samples \cite{lu2020nsganetv2}. Interestingly, as shown in Figure \ref{fig:predictors}, accuracy predictors can achieve acceptable mean absolute percentage error (MAPE) with far fewer training samples. Likewise, we have found that predictors perform well for latency and multiply-and-accumulates (MACs) objectives. We build on this insight that lightly trained predictors can offer a useful surrogate signal during search and combine this with the knowledge that GAs have been used for NAS applications with great success. 

Algorithm \ref{alg:linas} illustrates the LINAS flow where we first randomly sample the architecture search space to serve as the initial validation population. For each validation population, we measure each objective for the individuals and store the result. These results are combined with all previous validation population results and are used to train the objective predictors. For each iteration, we run a multi-objective genetic algorithm search (NSGA-II \cite{deb2002fast} in this work) using that iteration's trained predictors for a high number of generations (e.g., $>200$) to allow the algorithm to explore the predicted objective space sufficiently. This predictor-based GA search runs very quickly since no validation measurements occur. Finally, we select the most optimal population of diverse DNN architectures from the predictor-based GA search to add to the next validation population, which then informs the next round of predictor training. This cycle continues until the iteration count limit is met or an end-user decides a sufficient set of architectures has been discovered. We note that the LINAS approach can be applied with any single-, multi-, or many-objective evolutionary algorithm and generalizes to work with any fully-trained super-network framework. Additionally, it allows for the interchanging of GA tuning parameters (e.g. crossover, mutation, population), GA models, and predictor types for each iteration.

\begin{algorithm}[htb]
   \caption{Lightweight Iterative Neural Architecture Search}
   \label{alg:linas}
\begin{algorithmic}
   \STATE {\bfseries Input:}
   Objectives $f_m$, super-network with weights $\mathcal{W}$ and configurations $\Omega$, predictor model for each objective $Y_{m}$, LINAS population $P$ size $n$, number of LINAS iterations $I$, genetic algorithm $\mathcal{G}$ with the number of generations $J$.
   %\STATE \textcolor{gray}{// sample $n$ sub-networks for first population $P$}\\
   \STATE $P_{i=0} \leftarrow \{\omega_{n}\} \in \Omega$ \textcolor{gray}{// sample $n$ sub-networks for first population}\\ 
   \WHILE{$i++ < I$}
   %\STATE \textcolor{gray}{// measure objectives $f_m$, store data $D_{i,m}$} \\
   \STATE $D_{i,m} \leftarrow f_m(P_{i} \in \Omega; \mathcal{W})$ \textcolor{gray}{// measure $f_m$, store data $D_{i,m}$} \\
   \STATE $D_{all,m} \leftarrow D_{all,m} \cup D_{i,m}$
   \STATE $Y_{m,pred} \leftarrow Y_{m,train}(D_{all,m})$ \textcolor{gray}{// train predictors} 
   \WHILE{$j++ < J$} 
   \STATE $P_{\mathcal{G}_{j}} \leftarrow \mathcal{G}(Y_{m,pred}, j)$ \textcolor{gray}{// run ${\mathcal{G}}$ for $J$ generations}
   \ENDWHILE
   \STATE $P_{i} \leftarrow P_{\mathcal{G},best} \in P_{\mathcal{G}_{J}}$ \textcolor{gray}{// retrieve optimal population}
   %\STATE $i \leftarrow i+1$
   \ENDWHILE
   \STATE {\bfseries Output:} All LINAS validation populations $P_{I}$, GA predictor search results $P_{\mathcal{G}_{I,J}}$, and validation data $D_{all,m}$. 
   
\end{algorithmic}
\end{algorithm}

\section{Experiment}

\subsection{Experimental Setup}

We demonstrate the LINAS approach on the modalities of machine translation and image classification since they are representative of highly popular Transformer and convolutional DNN layer types respectively. Since our focus is on the DNN architecture search process, we do not re-train from scratch or fine-tune the optimal discovered architectures. For our experiments, we compare against validation-only measurements from a random search that uniformly samples the architecture space and NSGA-II for a well-known GA baseline. We leverage the pymoo \cite{pymoo} implementation of the NSGA-II algorithm and note that similar results can be achieved with AGE-MOEA \cite{agemoea}. 

Recent studies have looked at the performance for a wide range of predictor types for smaller DNN search spaces  \cite{white2021powerful}. For this work, the LINAS algorithm uses ridge and support vector machine regression (SVR) predictors with a one-hot encoding approach. In our studies, we found that these simpler methods converge more quickly, require fewer training examples, and require much less hyper-parameter optimization than multi-layer perceptrons (MLPs). The LINAS internal predictor-based GA uses NSGA-II with the same population, mutation, and crossover settings as the NSGA-II baseline. Table \ref{tab:search_settings} summarizes the DNN architectures, search space size, predictor types, and GA settings used in the experiments.

\begin{table}[htb]
\centering
\begin{tabular}{c|cc}
\hline \hline
\begin{tabular}[c]{@{}c@{}}Architecture\\ (Modality)\end{tabular} & \begin{tabular}[c]{@{}c@{}}Transformer\\ (Machine Translation)\end{tabular} & \begin{tabular}[c]{@{}c@{}}MobileNetV3\\ (Image Classification)\end{tabular} \\
\hline
Predictor & SVR w/ RBF kernel & Ridge \\
Search Space & $10^{15}$ & $10^{19}$ \\
Population & 50 & 50 \\
Crossover & 0.9 & 0.9 \\
Mutation & 0.02 & 0.02 \\
\hline
\end{tabular}
\caption{Experiment settings for both neural network architecture types. The predictor types apply only to the LINAS setup.}
\label{tab:search_settings}
\end{table}

\begin{figure}[tb]
    \centering
    \includegraphics[width=0.9\linewidth]{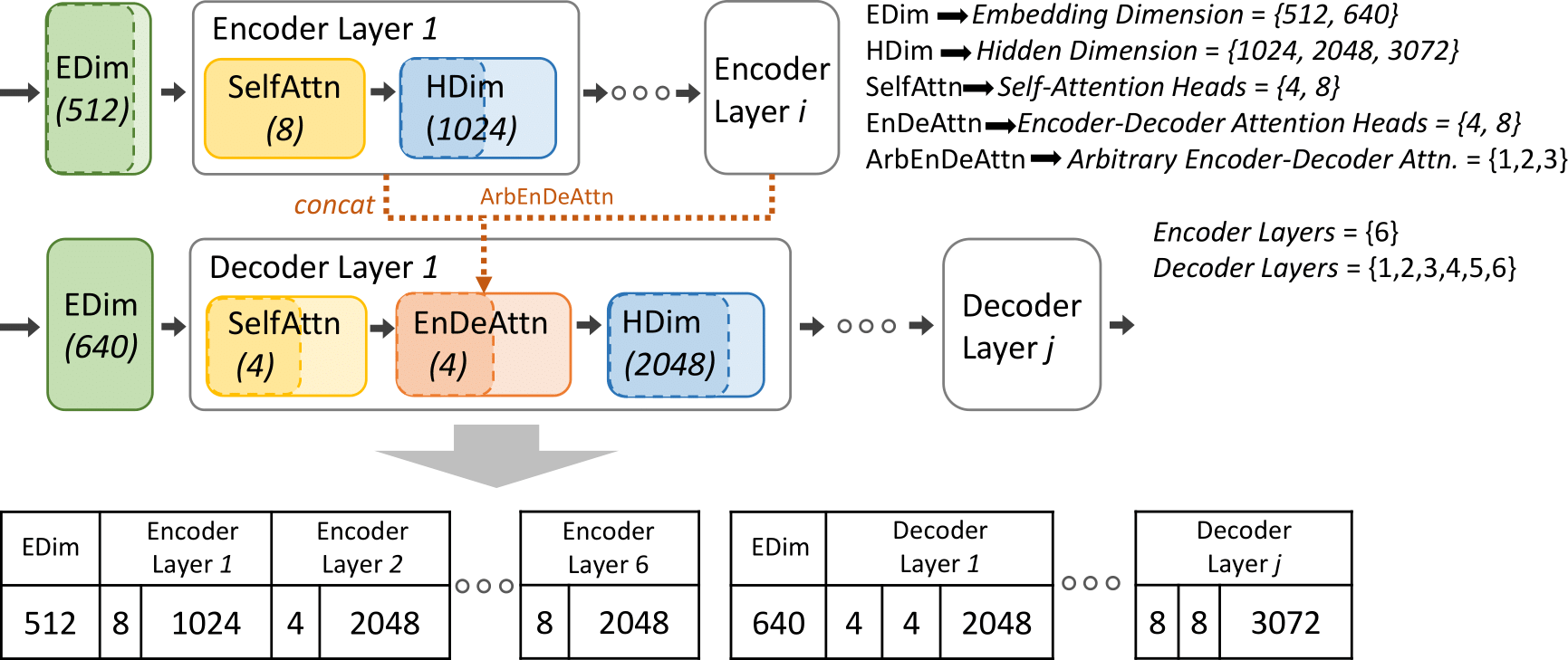}
    \caption{Encoding diagram for the Transformer search space which has 40 design variables.}
    \label{fig:hat_encoding}
\end{figure}

\begin{figure}[htb]
    \centering
    \includegraphics[width=0.9\linewidth]{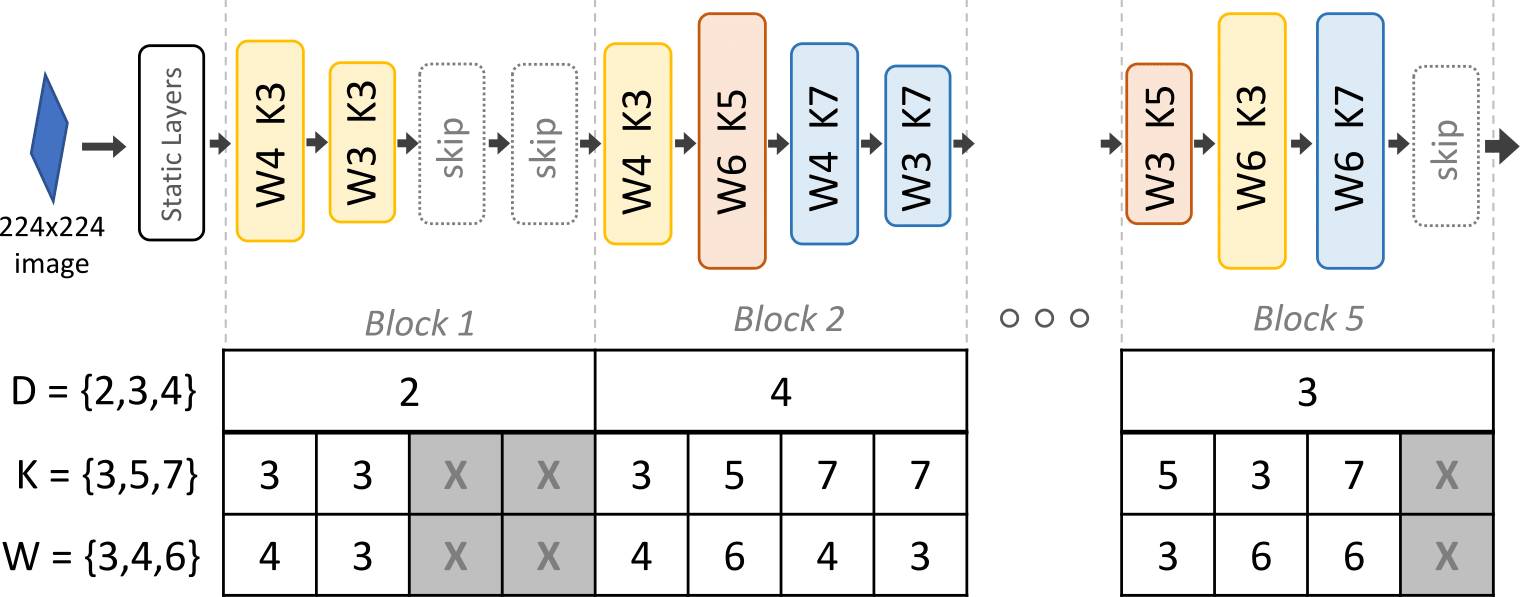}
    \caption{Encoding diagram of the MobileNetV3 search space which has 45 design variables.}
    \label{fig:mobilenetv3_encoding}
\end{figure}

Building a GA compatible encoding or representation of the architectural design variables is a critical step when applying GAs to NAS problems. We illustrate our encoding strategy in Figures \ref{fig:hat_encoding} and \ref{fig:mobilenetv3_encoding} that allows the GA operators (e.g., mutation, crossover, etc.) to run correctly and gives each design variable several integer options. For the machine translation modality, we run our experiments using the Transformer super-network. Our work is based on a recent and popular approach for machine translation \cite{wang2020hat} that samples the architecture encoder and decoder space during training to achieve a super-network model. We use a search space with an embedding dimension chosen from \{512, 640\} , hidden dimension from  \{1024, 2048, 3072\}, attention head number from \{4, 8\}, decoder layer number from \{1, 2, 3, 4, 5, 6\} and a constant encoder layer number \{6\}. Additionally, each decoder layer can attend to the last \{1,2,3\} encoder layers (arbitrary encoder-decoder attention). The first objective for the Transformer super-network is to maximize the bilingual evaluation understudy (BLEU) \cite{papinenibleu2002} score evaluated on the WMT 2014 En-De data set. For the BLEU score evaluation, we use a beam size of 5 and a length penalty of 0.6. 

For the computer vision modality, we perform an image classification task using a super-network based on the MobileNetV3 architecture \cite{cai2020onceforall}. In this framework, convolutional parameters such as block depth, channel width, and kernel size are used with a progressive shrinking approach during training. We allow for an layer depth chosen from \{2, 3, 4\}, an width expansion ratio chosen from \{3, 4, 6\}, and a kernel size chosen from \{3, 5, 7\} as shown in Figure \ref{fig:mobilenetv3_encoding}. The first objective is to maximize the top-1 accuracy using the ImageNet validation data set. The second objective for both modalities is to minimize latency and is chosen instead of MACs or DNN parameter counts since it is a more relevant hardware performance metric for real-life applications.

\subsection{Results}

\begin{figure}[tb]
    \centering
    \includegraphics[width=0.9\linewidth]{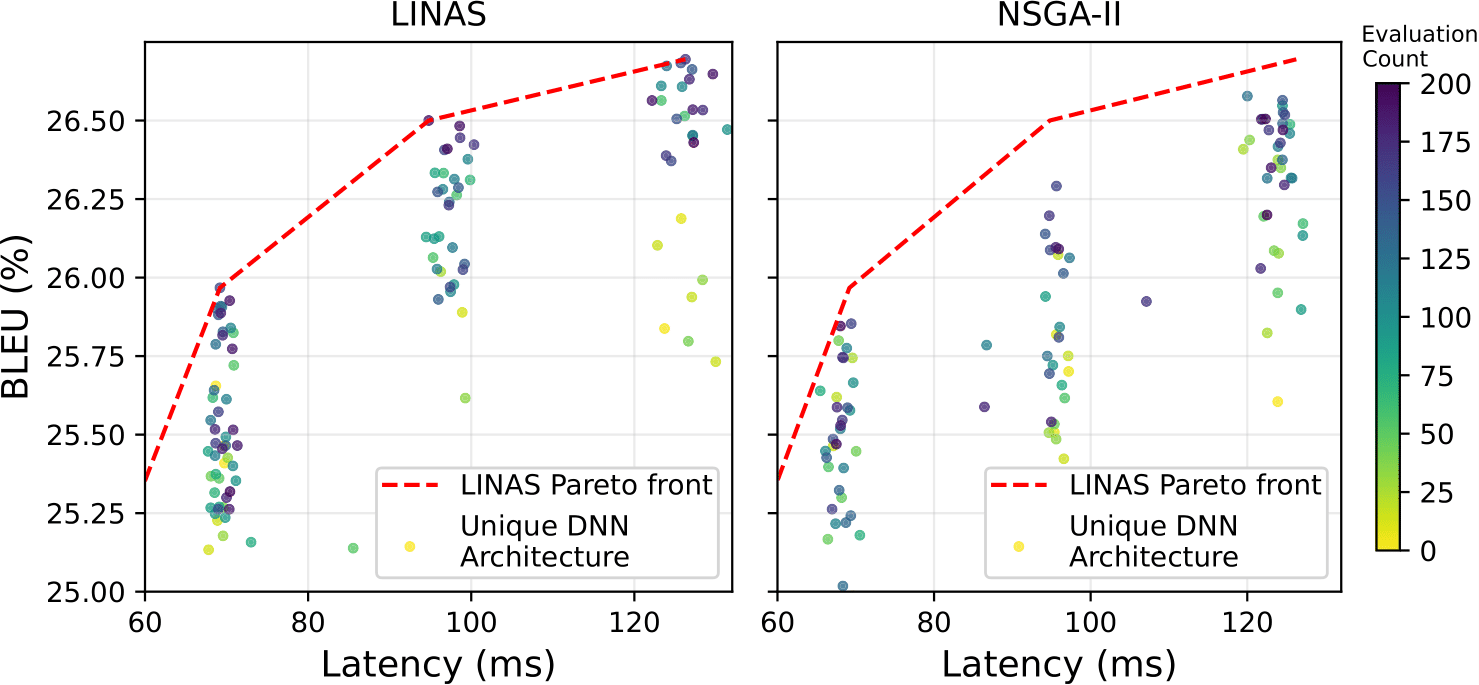}
    \caption{Comparison of LINAS and NSGA-II in the Transformer multi-objective search space for the same number of evaluations (validations) on a NVIDIA Titan-V GPU system.}
    \label{fig:transformer_scatter}
\end{figure}

\begin{figure}[tb]
    \centering
    \includegraphics[width=0.9\linewidth]{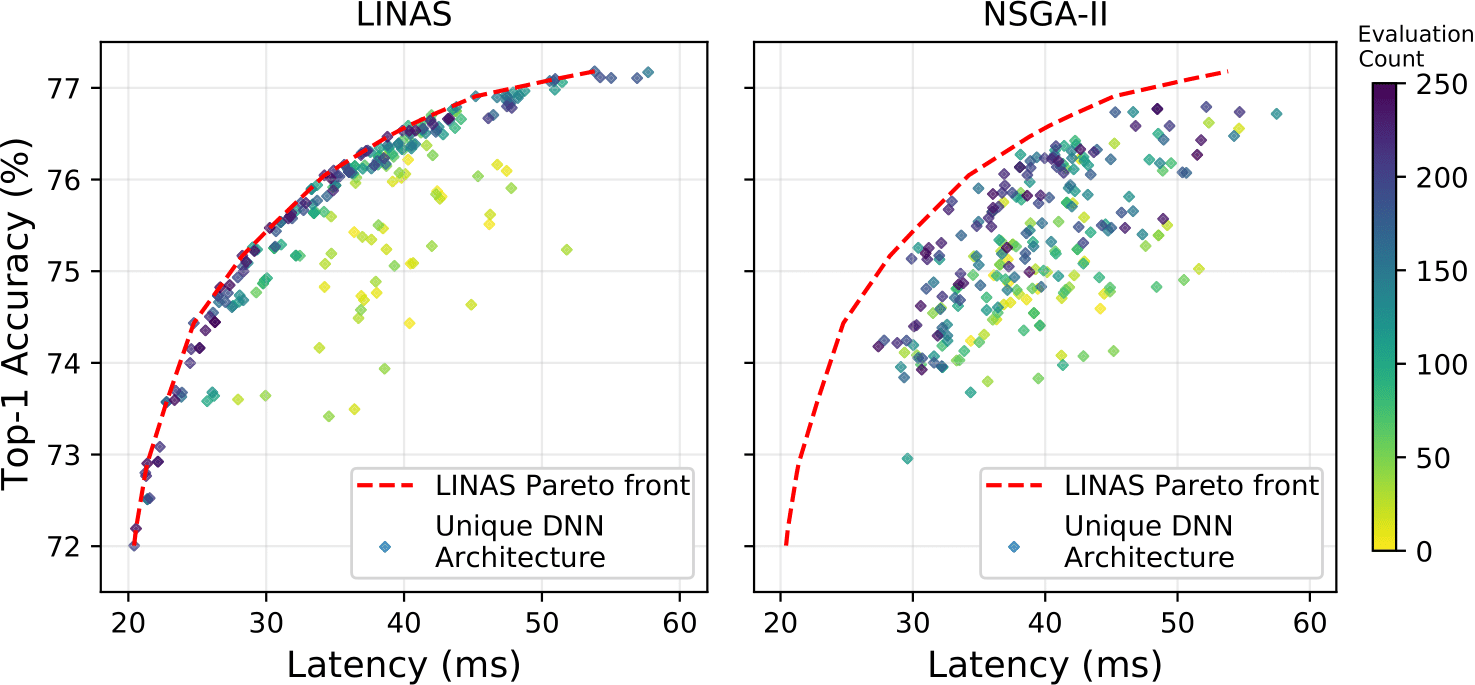}
    \caption{Comparison of LINAS and NSGA-II in the MobileNetV3 multi-objective search space for the same number of evaluations (validations) on a NVIDIA Titan-V GPU system.}
    \label{fig:mobilenetv3_scatter}
\end{figure}

The main purpose of our proposed LINAS approach is to reduce the total number of validation measurements required to find optimal DNN architectures in the multi-objective space for any modality (e.g., computer vision, machine translation). Specifically, for this work we want to efficiently discover architectures with optimal trade-offs in high accuracy/BLEU and low latency. Figures \ref{fig:transformer_scatter} and \ref{fig:mobilenetv3_scatter} illustrate the differences in how the NSGA-II baseline search evolves versus the proposed LINAS approach. While NSGA-II reliably progresses towards an optimal trade-off region, the LINAS results show how that exploration can be accelerated. An important note for discussion is that the Transformer architectural distributions in the BLEU and latency objective space fall into more discrete clusters that are a function of the decoder layers. Since the distribution of these models is both constrained in range and occurs closer to an optimal region, even if randomly sampled, the results look less differentiated than the MobileNetV3 results. 

\begin{figure}[tb]
    \centering
    \includegraphics[width=0.7\linewidth]{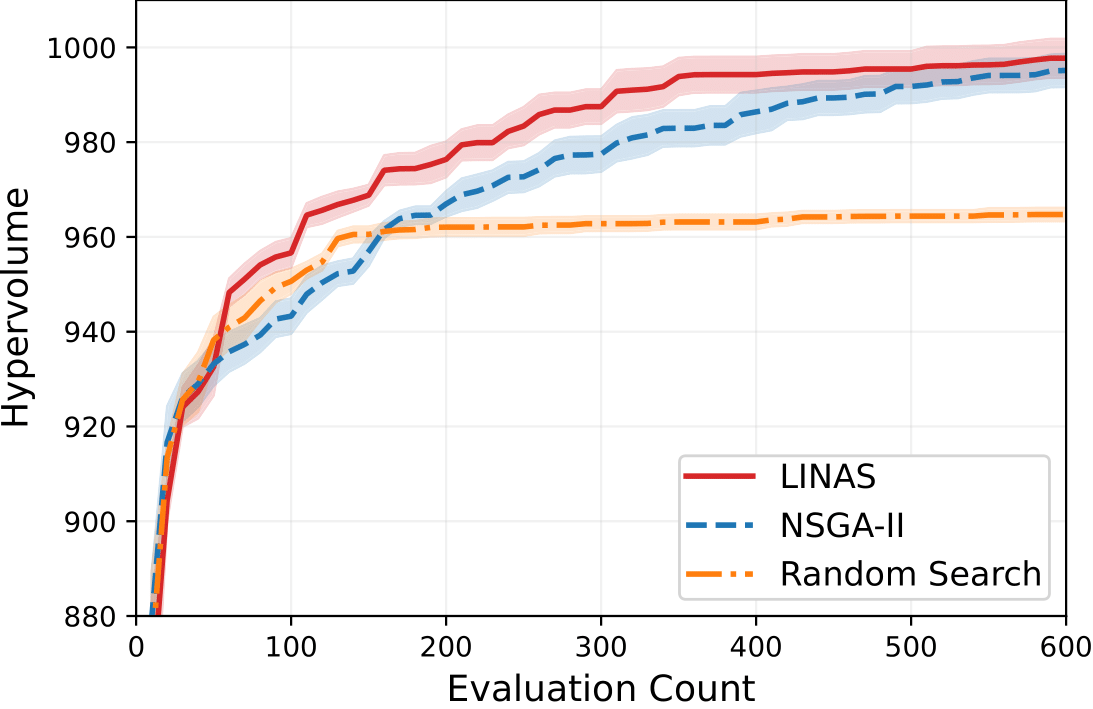}
    \caption{Hypervolume versus evaluation count in the machine translation Transformer search space (HV reference point BLEU=20, latency=200 ms). Shaded regions show the standard error for 5 trials.}
    \label{fig:hypervolume_hat}
\end{figure}

\begin{figure}[tb]
    \centering
    \includegraphics[width=0.7\linewidth]{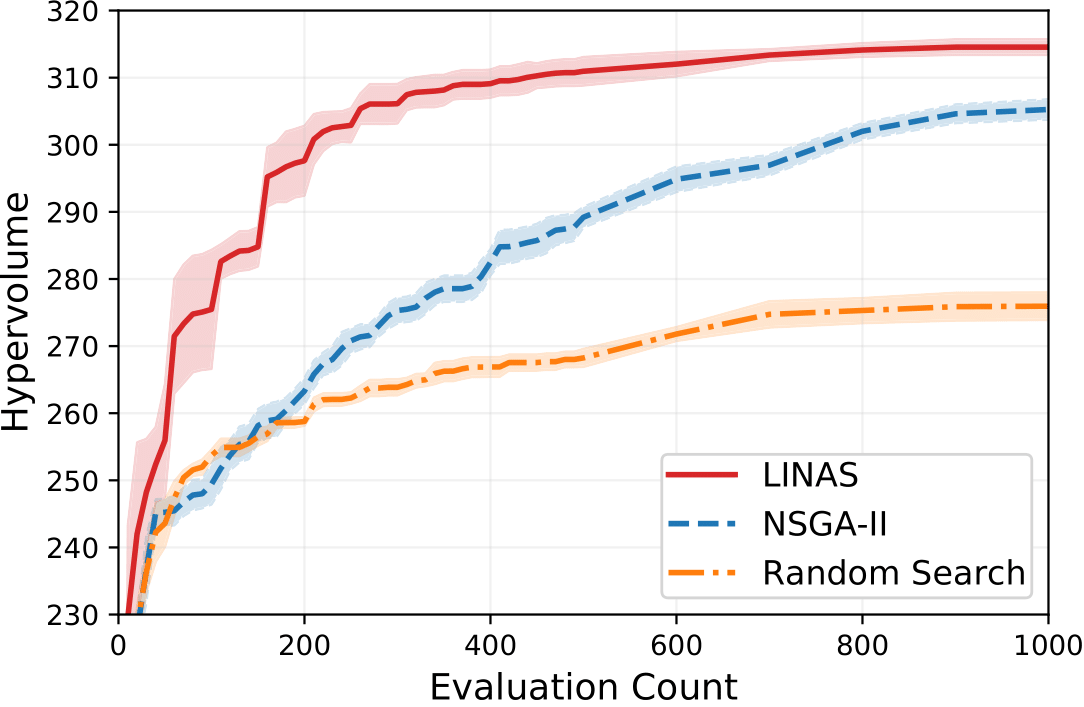}
    \caption{Hypervolume versus evaluation count for various search approaches in the image classification MobileNetV3 search space (HV reference point top-1=70\%, latency=70 ms). Shaded regions show the standard error for 5 trials.}
    \label{fig:hypervolume_mobilenetv3}
\end{figure}

Another way of visualizing the benefits of our approach is by evaluating the hypervolume (HV) versus evaluation counts. The hypervolume indicator \cite{zitzler1999multiobjective} offers a way to measure how well the Pareto front approximates the optimal solution. When measuring over two objectives, the hypervolume term represents the area of the Pareto front with respect to a reference point. Figures \ref{fig:hypervolume_hat} and \ref{fig:hypervolume_mobilenetv3} show the results of 5 trials for each algorithm with respect to the number of architecture evaluations (a.k.a. validation measurements). Each evaluation takes approximately 3 minutes in our setup meaning that 1000 evaluations would take about 50 GPU hours.

A key observation is how quickly LINAS accelerates to a better hypervolume versus the baseline NSGA-II search since each LINAS validation population after the first, represents the best predicted objective space information from the GA-predictor pairing. Depending on which region of the Pareto front is important, an end-user would be more likely to identify optimal architectures in fewer evaluations with LINAS. Given the constrained characteristics of the Transformer objective space and the higher MAPE in the BLEU predictor (see Figure \ref{fig:predictors}), the LINAS result is less differentiating for the Transformer results than in the MobileNetV3 case. 

% Running low on space
\begin{comment}
\begin{figure}[tb]
    \centering
    % \includegraphics{}
    \includegraphics[width=0.7\linewidth]{figures/hv_pareto_pic.png}
    \caption{Illustration of a two dimensional objective space with hypervolume and Pareto front point, where each point is representative of a DNN architecture.}
    \label{fig:hypervolume_metric}
\end{figure}
\end{comment}

\section{Conclusion}

The goal of the work was to demonstrate how GAs can be uniquely leveraged to accelerate multi-objective neural architecture search on the modalities of machine translation and image classification. The LINAS algorithm offers a modular framework that can easily be modified to fit a variety of NAS application domains. As NAS research continues to gain momentum, we highlight the need to continue to investigate the generalizability of NAS approaches in modalities outside of computer vision. Future work includes evaluating the use of proxy functions \cite{mellor2021neural} and advances in meta-learning \cite{lee2021help} to extend this type of algorithmic framework. 

\bibliographystyle{unsrt}  
\bibliography{references}

\end{document}